# A Novel Hybrid Ordinal Learning Model with Health Care Application

Lujia Wang, *Member, IEEE*, Hairong Wang, *Member, IEEE*, Yi Su, Fleming Lure, Jing Li, *Member, IEEE*, for the Alzheimer's Disease Neuroimaging Initiative*

*Abstract*— Ordinal learning (OL) is a type of machine learning models with broad utility in health care applications such as diagnosis of different grades of a disease (e.g., mild, modest, severe) and prediction of the speed of disease progression (e.g., very fast, fast, moderate, slow). This paper aims to tackle a situation when precisely labeled samples are limited in the training set due to cost or availability constraints, whereas there could be an abundance of samples with imprecise labels. We focus on imprecise labels that are intervals, i.e., one can know that a sample belongs to an interval of labels but cannot know which unique label it has. This situation is quite common in health care datasets due to limitations of the diagnostic instrument, sparse clinical visits, or/and patient dropout. Limited research has been done to develop OL models with imprecise/interval labels. We propose a new Hybrid Ordinal Learner (HOL) to integrate samples with both precise and interval labels to train a robust OL model. We also develop a tractable and efficient optimization algorithm to solve the HOL formulation. We compare HOL with several recently developed OL methods on four benchmarking datasets, which demonstrate the superior performance of HOL. Finally, we apply HOL to a real-world dataset for predicting the speed of progressing to Alzheimer's Disease (AD) for individuals with Mild Cognitive Impairment (MCI) based on a combination of multi-modality neuroimaging and demographic/clinical datasets. HOL achieves high accuracy in the prediction and outperforms existing methods. The capability of accurately predicting the speed of progression to AD for each individual with MCI has the potential for helping facilitate more individually-optimized interventional strategies.

*Note to Practitioners*— Machine learning (ML) algorithms have been widely adopted to support disease diagnosis and prognosis. In some situations, the outcome variable of interest is on an ordinal scale, i.e., it includes several classes with a natural order. For example, the variable of interest can be the grade of a disease as mild, moderate, or severe; or it can be the progression speed of a disease as very fast, fast, moderate, or slow. Ordinal learning (OL) is the type of ML algorithms for ordinal variable prediction. Most existing OL algorithms can only include samples with precise labels in training. However, it is common to have samples with imprecise/interval labels, i.e., we know that a sample belongs to a range of classes/labels but do not know which specific class/label it belongs to. This situation can happen due to a variety of different reasons such as use of less accurate diagnostic instrument under cost or availability constraints, sparse clinical assessment, and patient dropout. We propose a Hybrid Ordinal Learner (HOL) to integrate samples with both precise and interval labels to train a robust OL model. HOL is evaluated using four public benchmarking datasets and shows superior performance compared to existing methods. Also, we apply HOL to a real-world dataset for predicting the speed of progressing to Alzheimer's Disease (AD) for individuals with Mild Cognitive Impairment (MCI). MCI is the prodromal stage of AD. Individuals with MCI show noticeable signs of memory loss and cognitive declines, but these symptoms are not severe enough to interfere their independent living. HOL achieves high accuracy in predicting the speed of progressing to AD for each MCI subject (e.g., the speed of 'very fast', 'fast', 'moderate', or 'slow'), which could potentially help facilitate the development of more individually-optimized interventional strategies.

*Index Terms*—machine learning, ordinal learning, health care, imprecise labels

## I. INTRODUCTION

Machine learning (ML) algorithms have been widely adopted to support health care automation. Supervised learning is a type of ML algorithms, in which a model is trained to predict an outcome variable (y) based on features (x). Within supervised learning, the models can be future divided according to the type of the outcome variable. For categorical outcomes, classification models are used; for numerical outcomes, regression models can be appropriate. There is another type of models targeting the outcome variable that contains several classes but with a natural order, known as ordinal learning (OL) models. OL models have important utility in various health care applications such as diagnosis of different grades of a disease (e.g., mild, modest, severe) and prediction of the speed of disease progression (e.g., very fast, fast, moderate, slow).

Various OL models have been developed in the literature, such as support vector ordinal regression [1], ordinal forest [2], ordinal Gaussian Process [3]. A common limitation of these OL models is that they can only use precisely labeled samples in training, i.e., each sample must belong to one and only one class. However, in many applications, it is difficult to acquire a sufficient number of precisely labeled samples due to cost or availability constraints, whereas there could be an abundance of samples with imprecise/interval labels. For a sample of this kind,

L. Wang, H. Wang and J. Li are with H. Milton Stewart School of Industrial and Systems Engineering, Georgia Institute of Technology, Atlanta, GA 30332 USA (e-mail: {lwang724, hairongwhr}@gatch.edu, jing.li@isye.gatch.edu)

Y. Su is with Banner Alzheimer's Institute, Phoenix, AZ, 85006 USA (yi.su@bannerhealth.com).

Fleming Lure is with MS Technologies, Rockville, MD, 20850 USA (Fleming.Lure@mstechnologies.com).

*Data used in preparation of this article were obtained from the Alzheimer's Disease Neuroimaging Initiative (ADNI) database (adni.loni.usc.edu). As such, the investigators within the ADNI contributed to the design and implementation of ADNI and/or provided data but did not participate in analysis or writing of this report. A complete listing of ADNI investigators can be found at: http://adni.loni.usc.edu/wp-content/uploads/how_to_apply/ADNI_Acknowledgement_List.pdf

we can know that it belongs to an interval of labels, $[l, l+1, ..., l+k]$, but we cannot know which specific label the sample has. Next, we give some practical examples to illustrate this situation:

A motivating example in disease diagnosis: In determining the grade of a disease as mild, moderate, or severe, it may need expensive or invasive diagnostic approaches (e.g., imaging, biopsy) to determine the precise grade of the disease for a patient, whereas it may be relatively easier to find out if the disease belongs to an interval of grades such as mild-to-moderate using less expensive instrument such as symptom checklists. In this case, there can be limited samples with precise grades/labels in the training dataset but more samples with interval labels.

A motivating example in disease prognosis: In predicting the progression speed of a disease, the similar challenge exists with the training dataset. In the case study of this paper, we included a real-world application of using multi-modality neuroimaging data to predict the speed of progressing to Alzheimer's Disease (AD) for patients with Mild Cognitive Impairment (MCI) as very fast, fast, moderate, and slow. There are limited samples with precise progression speeds/labels due to sparse clinical assessments and patient dropout, whereas there are more samples with interval labels. In both the aforementioned diagnostic and prognostic scenarios, it is important to integrate samples with both precise and interval labels in order to train a robust OL model.

In this paper, we target a situation where one needs a model to predict the precise ordinal class label for each sample during deployment, such as the grade of a disease or speed of progression, but the dataset used to train this model comprises precisely-labeled and interval-labeled samples. Limited research has been done to develop OL models that can incorporate interval-labeled samples in training. Among the existing algorithms, Antoniuk et al. [4] was the first one to address this problem by introducing a specific V-shaped interval-insensitive loss under a linear ordinal classifier, which can be solved by a double-loop cutting plane solver. Later, Manwani and Chandra [5] proposed exact passive-aggressive (PA) online algorithms for learning to rank based on the interval insensitive loss under the linear ordinal classifier. Note that both papers only investigated linear ordinal classifiers. Manwani [6] extended the previous work into a non-linear ordinal classifier and proposed an online learning algorithm for parameter estimation. However, the estimated discriminative ordinal functions in the form of combinations of kernel functions are rough as the coefficients of kernels can only take integer values. On a parallel track, the problem of non-unique labels has been investigated in multi-class classification, and researchers have proposed various algorithms based on expectation-maximization [7][8], greedy heuristics [9], convex optimization [10], and maximum margin formulation [11]. However, these algorithms cannot be naturally extended for ordinal learning because they do not consider the order of classes/labels. Another related field investigates classification models with noisy labels, in which robust algorithms have been developed to tackle label noise/errors in training data [17-21]. It is important to note that noisy labels are different from interval/non-unique labels which is the focus of this paper. Noisy labels correspond to incorrect labels assigned to samples, whereas interval/non-unique labels involve true labels, but with uncertainty about which one within the interval is the correct label.

To fill the gap of existing research, we propose a new Hybrid Ordinal Learner (HOL) to integrate samples with precise and interval labels to train a robust OL model. The contributions of this paper are summarized as follows:

- **New model formulation**. We propose a non-linear model formulation of HOL. While most existing research in this area focuses on linear models, HOL provides the flexibility to address application domains where the data has complicated relationships. Also, this study identifies a sufficient and necessary condition for a proper loss function when the training samples include interval-labeled samples. We further propose a general form of the loss function that satisfies this condition. This not only helps us design the loss function of HOL in this paper, but also provides a general framework that allows more work to be done for developing OL models with interval-labeled samples by other interested researchers.
- **Tractable and efficient optimization algorithm**. The challenge for non-linear model formulation is loss of tractability and efficiency in model estimation. To address this challenge, we propose a novel conversion method that converts the HOL formulation into an equivalent formulation of learning a set of binary classifiers with coupled parameters. Because binary classification has been more studied in the literature, this conversion allows us to borrow ideas from binary classification. Furthermore, we propose convex surrogates for the loss function in the converted formulation, which allows the optimization to be solved using efficient convex solvers.
- **Benchmarking experiments**. We compare the performance of HOL with several recently developed OL models that can accommodate interval labels on four benchmarking datasets. HOL shows better performance.
- **Contribution to early prediction of AD.** We apply HOL to an application of predicting the speed of progressing to AD for individuals with MCI—an early stage before dementia—based on multi-modality neuroimaging and demographic/clinical datasets. Accurate prediction of the speed of progressing to AD for each MCI subject as 'very fast', 'fast', 'moderate', or 'slow' is important for facilitating the development of more individually-optimized interventional strategies. HOL achieves high accuracy in the prediction and outperforms competing methods.

## II. RELATED WORKS

### A. Ordinal Learning (OL) methods

The existing OL models mainly fall into four categories. One category transforms the ordinal scales into numeric values, which converts the problem into a regression problem [12]. This is an over-simplification for the metric distance between different ordinal labels. The second category of models decomposes the ordinal problem into classification tasks [13]. However, this does not take full advantage of ranking information of ordinal classes. The third category of algorithms fits a regression with a set of thresholds to identify the ranking responses. Shashua et. al [14] generalized support vector methods to ordinal regression via defining a set of thresholds and finding the corresponding parallel hyperplanes, i.e., sandwiching ordered classes. Chu et. al [1] modified the support vector ordinal regression through imposing ordinal inequality constraints on the thresholds explicitly and implicitly to guarantee that the natural ordinal inequalities on the thresholds hold at the solution. Chu et. al [3] also proposed ordinal Gaussian Process, where a set of thresholds were introduced into likelihood functions of ordinal variables to enforce the ordinal constraints. The fourth category of algorithms is to formulate a large, augmented classification problem by considering the ordinal information. Har-Peled et al. [15] introduced a constrained classification framework that was able to solve ranking and multilabel classification problems. Herbrich et al. [16] showed that each ordinal regression problem can be converted to a preference learning problem based on pairs of objects, which built the connection to classification approaches for ordinal learning. However, all the aforementioned algorithms cannot incorporate interval-labeled samples in training.

### B. OL and other machine learning methods with interval/non-unique/noisy labels

Limited research has been done to develop OL models with interval labels. Antoniuk et al. [4] was the first one to address this problem by introducing a specific V-shaped interval-insensitive loss under a linear ordinal classifier, which can be solve by a double-loop cutting plane solver. Later, Manwani and Chandra [5] proposed exact passive-aggressive (PA) online algorithms for learning to rank based on the interval insensitive loss under the linear ordinal classifier. Note that both papers only investigated linear ordinal classifiers. Manwani [6] extended the previous work into a non-linear ordinal classifier and proposed an online learning algorithm for parameter estimation. However, this algorithm is rough, allowing only integer coefficients for the representative discriminative functions.

Another related field is multi-class classification in which research has been conducted to address the lack of unique labels for some samples. A sample of this kind is associated with a set of potential labels, but it is unknown which specific label it has. This is a similar situation to the interval label in OL, but the concept of "interval" does not apply here because there is no order of the labels/classes in multi-class classification. To address non-unique labels in multi-class classification, some researchers proposed to use expectation-maximization based algorithms to fill in the 'missing' labels and estimate the parameters [7][8]. Hullermeier and Beringer [9] extended three classification methods, Nearest Neighbor Classification, Decision Tree Induction, Rule Induction to address non-unique labels based on greedy heuristics. Cour et al. [10] proposed a convex learning formulation by discriminating the average output from all candidate labels against the outputs from non-candidate labels. Yu and Zhang [11] developed a maximum margin formulation to directly optimize the margin between the ground truth label and all other labels. However, all these existing algorithms focus on multi-class classification problems in which the class labels do not have an order. They are not suitable for ordinal learning, which is the focus of this paper.

In multi-class classification, some methods have been developed to handle noisy labels, where some training samples may be labeled incorrectly. For instance, Krawczyk et al. [17] utilized fuzzy one-class classifiers assigned to specific classes to capture distinguishing characteristics and mitigate label noise. Wang et al. [18] employed an importance reweighting strategy applicable across various loss functions and classification settings, providing an upper bound on the proportion of randomly labeled examples in noise-free scenarios. Wu et al. [19] introduced a method using a noise transition matrix to learn from solely noisy-similarity-labeled data. Villacampa-Calvo et al. [20] developed Gaussian Process classifiers with the capability of incorporating prior information of known noise levels. Ding et al. [21] proposed decomposing the loss function and transforming the recovery problem into a centroid estimation problem for unbiased risk estimation in multi-class learning. However, these methods are devised for multi-class classification problems and therefore not directly applicable to ordinal classes. Moreover, noisy labels differ from interval/non-unique labels which are the specific focus of this paper. Specifically, noisy labels refer to erroneous labels assigned to samples, whereas interval/non-unique labels contain the true labels but with uncertainty about which one within the interval is the correct label.

### C. Prediction of MCI progression to AD using multi-modality neuroimaging data and machine learning

This work is motivated by the application domain of using multi-modality neuroimaging datasets for early prediction of the progression to AD for individuals with MCI. We briefly review the existing research in this area.

Neuroimaging has shown great promise for early prediction of AD [22], [23]. Multi-modality neuroimaging datasets provide complementary measures for different aspects of the brain affected by the disease [24]. Past research has shown improved prediction capability of combining multi-modality neuroimaging datasets compared to using a single modality [25], [26]. Two commonly used modalities are MRI and PET, which measure structure and function of the brain, respectively. Most existing research combining MRI and PET for predicting MCI progression to AD formulates the prediction task as a binary classification problem, which classifies an individual with MCI as a converter if the individual progresses to AD by a pre-defined timeframe, and as a non-converter otherwise. Various machine learning methods have been proposed. For example, Shen et al. [27] integrated PET with MRI through a sparse regression method for MCI conversion prediction. Zhang et al. [28] utilized an attention mechanism in the proposed deep multi-modal fusion network to extract discriminative features from MRI and PET for MCI classification. Zhou et al. [29] proposed a three-stage multi-modality feature learning framework, in which independent latent representations for each modality, joint latent features for pairs of modality combinations, and diagnostic labels were learned stage by stage. Zhou et al. [30] also proposed a latent feature representation learning framework which learned common latent representations through samples with all modalities and modality-specific latent representations through samples with specific modalities. Zhou et al. [31] further mapped multi-modality neuroimaging data into learned latent representations, and the final classification results were obtained by an ensemble strategy for diversified support vector machine (SVM) classifiers which were trained to project the latent representations into the label space. Zhu et al. [32] developed a rank minimization multiple-kernel learning model which imposed low-rank constraint on the regression coefficient matrix of all modalities and adaptively measured the contribution of each sample by self-paced learning. Liu et al. [33] introduced a view-aligned hypergraph learning (VAHL) method using incomplete multi-modality data, which partitioned data into views according to the availability of different modalities and learned sparse representation to construct a hypergraph in each view. Liu et al. [34] proposed an Incomplete-Multimodality Transfer Learning (IMTL) model by building predictive models based on different sub-cohorts of samples with same missing modalities and combining the model estimation processes to allow for transfer learning.

However, all these existing methods are binary classifiers, which cannot predict the different speeds of progressing to AD for each MCI subject—a capability provided by an OL model. To our best knowledge, little research has been done to predict the speed of MCI progression to AD using OL models, especially by leveraging datasets with imprecise/interval labels to build robust models.

*D. Gaps in the existing research*

As reviewed in the previous sections, there are gaps in both the machine learning field and the application domain. In machine learning, most existing OL models cannot integrate interval-labeled samples in training. There is only a handful of recent studies developing OL models with interval labels as mentioned previously. However, these studies either focus on linear models or lack rigorous model estimation procedures. In a related field of multi-class classification, some methods have been developed to incorporate samples with non-unique labels in training. However, these methods cannot be directly used to build an OL model because they do not account for the intrinsic order of the class labels. Some other methods have been developed to tackle noisy labels. However, these approaches have a different focus from ours. Noisy labels are incorrect labels assigned to samples, whereas interval/non-unique labels contain the true labels but with uncertainty about which one within the interval is the correct label. Similarly, in the application domain, most existing studies treat the problem of predicting MCI progression to AD as a binary classification problem. There is a lack of studies for progression speed prediction via training robust OL models with interval labels.

Therefore, in this paper, we aim to fill the aforementioned gaps by developing a novel HOL method, which can integrate samples with precise and interval labels to train a robust OL model. We demonstrate the superior performance of HOL in predicting the speed of progression to AD for individuals with MCI.

### III. PROPOSED HYBRID ORDINAL LEARNER (HOL)

*A. Notations*

Let $\mathcal{X} = \mathbb{R}^d$ be the $d$-dimensional feature space and $\mathcal{Y} = 1, \ldots, K$ be the label space. Different from multi-class classification, the consecutive integers in $1, \ldots, K$ follow an order. In this paper, we use 'labels' and 'classes' interchangeable to refer to the integers contained in $\mathcal{Y}$.

A typical OL model consists of a set of ranking functions, $f_k, k = 1, \ldots, K-1$, which satisfy the constraint of $f_1 \leq \cdots \leq f_{K-1}$. To predict the label for a sample $\mathbf{x}$, one can compute the outputs from the ranking functions for this sample, $f_1(\mathbf{x}), \ldots, f_{K-1}(\mathbf{x})$. Then, the predicted label can be obtained by

$$J(\mathbf{x}) = 1 + \sum_{k=1}^{K-1} I(f_k(\mathbf{x}) < 0), \quad (1)$$

where $I(\cdot)$ is an indicator function. The meaning of (1) is that the predicted label, $J(\mathbf{x})$, is the number of negative ranking functions in the sequence plus one. For example, if all ranking functions are non-negative, the predicted label is 1; if all ranking functions are negative, the predicted label is $K$; if the first $k$ ($1 \leq k < K-1$) ranking functions are negative and the remaining ones are non-negative, the predicted label is $k + 1$.

Furthermore, the ranking functions can be decomposed into a common function and class-specific intercepts, $f_k(\mathbf{x}) = h(\mathbf{x}) + b_k$ with $b_1 \leq \cdots \leq b_{K-1}$ [1]. $h(\mathbf{x}) = \boldsymbol{\eta}^T \boldsymbol{\phi}(\mathbf{x})$, where $\phi$ includes transformations of the feature vector $\mathbf{x}$ and $\boldsymbol{\eta}$ contains the combination coefficients. Depending on the form of $\phi$, the OL model can be linear or non-linear. A training dataset is needed to learn the parameters such as $\boldsymbol{\eta}, b_1, \ldots, b_{K-1}$. For a complete list of mathematical notations used in the remainder of this paper, please see Appendix. D.

*B. Mathematical formulation of HOL*

Consider a training dataset of $n$ samples, $(\mathbf{x}_i, \mathcal{F}_i), i = 1, \ldots, n$. In the conventional OL setting, every training sample must have one and only one label. HOL allows the training set to include samples with interval labels, i.e., $\mathcal{F}_i = [Y_i^l, Y_i^r] \subseteq \mathcal{Y}$. For example, a sample may have an interval label of $[2,4]$, meaning that the sample can be from class 2, 3, or 4, but we do not know which precise class it is from. HOL can incorporate samples with both interval and precise labels. When $Y_i^l < Y_i^r$, $\mathcal{F}_i$ denotes an interval label. When $Y_i^l = Y_i^r$, $\mathcal{F}_i$ denotes a precise label.

The goal of HOL is to learn an OL model based on a training set with the aforementioned characteristics. This can be formulated as the following optimization problem:

$$\min_{\mathbf{f}=(f_1,\ldots,f_{K-1})} \sum_{i=1}^{n} L(J(\mathbf{x}_i),\mathcal{F}_i) + \mu||\mathbf{f}||_{\mathcal{H}},$$
$$\text{s.t. } f_1 \leq \cdots \leq f_{K-1}, \quad (2)$$

where $L$ is a loss function defined on the training set that will be discussed later, $||\cdot||_{\mathcal{H}}$ is a norm in a metric space $\mathcal{H}$ to regularize the complexity of the ranking functions, and $\mu$ controls the trade-off between the loss and model complexity.

Because $\mathcal{F}_i$ can be an interval, commonly used loss functions for supervised learning models are not applicable. In what follows, we first identify the sufficient and necessary condition for a proper loss function of HOL (Definition 1). Then, we propose a general form of the loss function that satisfies the condition (Proposition 1). Finally, we proposed two specific loss functions used in this paper (Proposition 2 and 3). For notation simplicity, denote $L(J(\mathbf{x}_i),\mathcal{F}_i)$ by $L(J_i,\mathcal{F}_i)$ hereafter.

**Definition 1**: A proper loss function for HOL, $L(J_i,\mathcal{F}_i)$, must satisfy two conditions:
(a) When the predicted label $J_i$ falls within the true label interval $\mathcal{F}_i = [Y_i^l, Y_i^r]$, the prediction is considered correct and the loss is zero. That is,

$$L(J_i,\mathcal{F}_i) = 0 \text{ if } J_i \in [Y_i^l, Y_i^r].$$

(b) When $J_i$ falls outside $\mathcal{F}_i$, the loss should not decrease as $J_i$ is farther away from $\mathcal{F}_i$. That is:

$$\begin{cases} L(J_i,\mathcal{F}_i) \leq L(J_i - 1, \mathcal{F}_i) & \text{if } J_i < Y_i^l \\ L(J_i,\mathcal{F}_i) \leq L(J_i + 1, \mathcal{F}_i) & \text{if } J_i > Y_i^r \end{cases}.$$

For example, if the true label interval of a sample is [2,4], the loss should be zero if the predicted label is 2, 3, or 4, according to condition (a). An example for condition (b) is the following: consider two samples $i$ and $j$ with the same true label interval [2,4], whose predicted labels are 5 and 6, respectively. Then, the loss for sample $i$ should not be higher than that for sample $j$. That is, the loss for sample $i$ should be lower than that for sample $j$, if we want to penalize more for predictions that are farther away from the true interval; or the losses of the two samples should be equal, if we want to penalize predictions in the same way as long as they are not in the true interval.

**Proposition 1 (general form for the loss in HOL)**: A loss function $L(J_i,\mathcal{F}_i)$ that satisfies the conditions in Definition 1 can be expressed as:

$$L(J_i,\mathcal{F}_i) = \begin{cases} 0 & if J_i \in [Y_i^l, Y_i^r] \\ \sum_{k=J_i}^{Y_i^l-1} w_{k,i} & \text{if } J_i < Y_i^l \\ \sum_{k=Y_i^r}^{J_i-1} w_{k,i} & \text{if } J_i > Y_i^r \end{cases}, \quad (3)$$

for any $w_{k,i} \geq 0$.

*Proof:* For notation simplicity, the subscript $i$ is omitted in the following derivation. According to the Definition 1, the loss function $L(J,\mathcal{F})$ can be decomposed as

$$L(J,\mathcal{F}) = \begin{cases} 0 & \text{if } J \in [Y^l, Y^r] \\ L(J,Y^l) & \text{if } J < Y^l \\ L(J,Y^r) & \text{if } J > Y^r \end{cases}$$

$$= \begin{cases} 0 & \text{if } J \in [Y^l, Y^r] \\ \sum_{k=J}^{Y^l-1}(L(k,Y^l) - L(k+1,Y^l)) + L(Y^l,Y^l) & \text{if } J < Y^l \\ \sum_{k=Y^r}^{J-1}(L(k+1,Y^r) - L(k,Y^r)) + L(Y^r,Y^r) & \text{if } J > Y^r \end{cases}.$$

According to condition (a) in Definition 1, $L(Y^l,Y^l) = 0$ and $L(Y^r,Y^r) = 0$. Define $w_k$ by

$$w_k = \begin{cases} 0 & \text{if } k \in [Y^l, Y^r] \\ L(k,Y^l) - L(k+1,Y^l) & \text{if } k < Y^l \\ L(k+1,Y^r) - L(k,Y^r) & \text{if } k > Y^r \end{cases}.$$

According to condition (b) in Definition 1, we can know that $w_k \geq 0$ for $k < Y^l$ and $w_k \geq 0$ for $k > Y^r$. That is $w_k \geq 0$ for all $k$. Then we have

$$L(J,\mathcal{F}) = \begin{cases} 0 & \text{if } J \in [Y^l, Y^r] \\ \sum_{k=J}^{Y^l-1} w_k & \text{if } J < Y^l \\ \sum_{k=Y^r}^{J-1} w_k & \text{if } J > Y^r \end{cases}. \blacksquare$$

In theory, any loss function that can be expressed by (3) is applicable for HOL. In this paper, we focus on two specific forms of the loss function for computational ease, i.e., the Mean Absolute Error (MAE) loss and the 0/1 loss, which are given as follows:

**Proposition 2 (MAE loss in HOL)**: The MAE loss is the minimum distance between the predicted label $J_i$ and the true interval $\mathcal{F}_i = [Y_i^l, Y_i^r]$, i.e.,

$$L_{MAE}(J_i,\mathcal{F}_i) = \min_{Y \in \mathcal{F}_i}|Y - J_i| = \begin{cases} 0 & if J_i \in [Y_i^l, Y_i^r] \\ Y_i^l - J_i & \text{if } J_i < Y_i^l \\ J_i - Y_i^r & \text{if } J_i > Y_i^r \end{cases}. \quad (4)$$

**Proposition 3 (0/1 loss in HOL)**: A loss of '1' is incurred if the predicted label $J_i$ falls outside the true interval $\mathcal{F}_i$, and the loss is '0' otherwise, i.e.,

$$L_{0/1}(J_i,\mathcal{F}_i) = I(J_i \notin \mathcal{F}_i). \quad (5)$$

It is straightforward to prove that (4) and (5) satisfy the general form in (3). Here we provide an example to illustrate these loss functions. Consider a sample with true label interval [2,4]. Under the MAE loss, if the predicted label is 5, the loss is 5-4=1; if the predicted label is 6, the loss is 6-4=2. Under the 0/1 loss, both predictions will incur the same loss of 1.

## C. Conversion of the HOL Optimization

It is difficult to solve the HOL optimization in (2) especially when the ranking functions $f_1, ..., f_{K-1}$ are non-linear. This is because the ranking functions are embedded in the loss functions in a complicated form, which makes the optimization intractable. To tackle this challenge, we propose a conversion method that converts the original HOL optimization into an equivalent formulation of learning $K - 1$ binary classifiers with coupled parameters. Because binary classification has been more studied in the literature, this conversion allows us to borrow ideas from binary classification to effectively and efficiently solve the HOL optimization.

Specifically, consider each ranking function $f_k$ to be a binary classifier: if $f_k(\mathbf{x}) > 0$, $\mathbf{x}$ is classified to the interval $[1, k]$; otherwise, $\mathbf{x}$ is classified to the interval $[k + 1, K]$. To train $f_k$, we use a subset of training samples whose label $\mathcal{F}_i$ is included in $[1, k]$ or $[k + 1, K]$. This is a subset of the whole training set excluding samples whose label interval includes $k$. Denote this subset by:

$$D_k = \{(\mathbf{x}_i, \mathcal{F}_i) | \mathcal{F}_i \subseteq [1, k] \text{ or } \mathcal{F}_i \subseteq [k + 1, K]; i = 1, ..., n\}.$$

Next, we can define the loss function of training $f_k$ as:
$$\sum_{i \in D_k} w_{k,i} I(Z_{k,i} f_k(\mathbf{x}_i) < 0),$$

where $Z_{k,i} = 1, -1, \text{ or } 0$ corresponds to $\mathcal{F}_i \subseteq [1, k], \mathcal{F}_i \subseteq [k + 1, K], \text{ or } k \in \mathcal{F}_i$, respectively. A loss of '1' is incurred for a sample if the predicted and true classes of the sample do not agree. $w_{k,i}$ has been defined in (2) and is used to weight different samples in the loss function. Finally, we can sum up the loss of each $f_k$ and get the total loss for training the $K - 1$ binary classifiers simultaneously, i.e.,

$$\mathcal{B}(f, Z) \triangleq \sum_{k=1}^{K-1} \sum_{i \in D_k} w_{k,i} I(Z_{k,i} f_k(\mathbf{x}_i) < 0). \quad (6)$$

Theorem 1 proves that $\mathcal{B}(f, Z)$ is equivalent to the HOL loss in (2). The proof of Theorem 1 can be found in Appendix. A.

**Theorem 1:** Let $L(J, \mathcal{F}) \triangleq \sum_{i=1}^{n} L(J(\mathbf{x}_i), \mathcal{F}_i)$ denote the HOL loss in (2), where the sample-wise loss is in the general form in Proposition 1. $\mathcal{B}(f, Z)$ is the total loss of $K - 1$ binary classifiers based on $f_1, ..., f_{K-1}$, as defined in (6). Then, $L(J, \mathcal{F}) = \mathcal{B}(f, Z)$.

Based on Theorem 1, we can convert the HOL optimization in (2) into an equivalent form as:

$$\min_{\mathbf{f}=(f_1,...,f_{K-1})} \sum_{k=1}^{K-1} \sum_{i \in D_k} w_{k,i} I(Z_{k,i} f_k(\mathbf{x}_i) < 0) + \mu ||\mathbf{f}||_{\mathcal{H}},$$
$$\text{s.t. } f_1 \leq \cdots \leq f_{K-1}. \quad (7)$$

To solve this optimization problem is to train $K - 1$ binary classifiers with coupled parameters in $f_1, ..., f_{K-1}$, which is more tractable than solving the original optimization. In the next section, we will present the algorithms to solve the optimization in (7).

## D. Algorithm for solving the HOL optimization

To solve the optimization in (2), we first propose to use the hinge loss as a surrogate for the indicator function in (7) to make the optimization more tractable and efficient to solve. The hinge loss is a convex upper bound of the indicator function. Using the hinge loss and spelling out the ranking functions as $f_k(\mathbf{x}_i) = \boldsymbol{\eta}^T \boldsymbol{\phi}(\mathbf{x}_i) + b_k$, (2) becomes:

$$\min_{\boldsymbol{\eta}, b_1, ..., b_{K-1}} \sum_{k=1}^{K-1} \sum_{i \in D_k} w_{k,i} \max\left(0, 1 - Z_{k,i}(\boldsymbol{\eta}^T \boldsymbol{\phi}(\mathbf{x}_i) + b_k)\right) +$$
$$\mu ||\boldsymbol{\eta}||_{\mathcal{H}},$$
$$\text{s.t. } b_1 \leq \cdots \leq b_{K-1}. \quad (8)$$

Next, we will present the algorithms for solving (8) under the MAE loss and 0/1 loss, respectively:

### D.1 Algorithm for solving the HOL optimization under the MAE loss

It can be shown that $w_{k,i} = 1$ if $k \leq Y_i^l - 1$ or $k \geq Y_i^r$ under the MAE loss. Furthermore, Theorem 2 indicates that the constraint in (8) is automatically satisfied with the MAE loss, so that we can solve an unconstrained optimization problem. The proof of Theorem 2 can be found in Appendix. B.

**Theorem 2**: Under the MAE loss, the solution of the constrained optimization in (8) is the same as that of the optimization without the constraint. (See proof in Appendix. B)

Thus, the optimization in (8) under the MAE loss becomes:

$$\min_{\boldsymbol{\eta}, b_1, ..., b_{K-1}} \sum_{k=1}^{K-1} \sum_{i \in D_k} \max\left(0, 1 - Z_{k,i}(\boldsymbol{\eta}^T \boldsymbol{\phi}(\mathbf{x}_i) + b_k)\right) +$$
$$\mu ||\boldsymbol{\eta}||_{\mathcal{H}}. \quad (9)$$

To solve (9), we borrow the max-margin concept from SVM [35] and represent (9) into an equivalent form by introducing slack variables $\xi_i^k$ and a tuning parameter $\lambda$, i.e.,

$$\min_{\boldsymbol{\eta}, b_1, ..., b_{K-1}, \xi} \frac{1}{2} ||\boldsymbol{\eta}||^2 + \lambda \sum_{k=1}^{K-1} \sum_{i \in D_k} \xi_i^k$$
$$\text{s.t. } Z_{k,i}(\boldsymbol{\eta}^T \boldsymbol{\phi}(\mathbf{x}_i) + b_k) \geq 1 - \xi_i^k;$$
$$\xi_i^k \geq 0;$$
$$i \in D_k, k = 1, ..., K - 1. \quad (10)$$

To efficiently solve the optimization in (10), we derived the dual form of the primal problem, which is summarized in Theorem 3. The proof can be found in Appendix. C.

**Theorem 3:** Let $\mathbf{Z}$ denote a diagonal matrix, $\mathbf{Z} = diag\left(\overbrace{Z_{1,1}, ..., Z_{1,n}}^{n}, ..., \overbrace{Z_{K,1}, ..., Z_{K,n}}^{n}\right)$. Let $\mathbf{C}$ denote a covariance matrix with $C_{ij} = \phi(\mathbf{x}_i)^T \phi(\mathbf{x}_j) = k(\mathbf{x}_i, \mathbf{x}_j)$, which can be computed by a kernel function $k(\cdot)$ defined on the feature space. Then, the dual form of the primal HOL optimization in (10) is:

$$\min_{\gamma} \frac{1}{2}\gamma^T ZCZ\gamma - \sum_{k=1}^{K-1}\sum_{i\in D_k} \alpha_i^k,$$

$$\text{s.t. } \sum_{i\in D_k} \alpha_i^k Z_{k,i} = 0, k = 1, \dots, K-1;$$

$$0 \leq \alpha_i^k \leq \lambda, i \in D_k, k = 1, \dots, K-1, \quad (11)$$

where $\gamma = (\{\alpha_{i\in D_1}^1\}, \dots, \{\alpha_{i\in D_{K-1}}^{K-1}\})^T$ are the Lagrange multipliers and $\lambda$ is a tuning parameter.

The dual form of the optimization in (11) is convex and can be solved by a quadratic programing solver, such as CPLEX. When the sample size is large, the sequential minimal optimization (SMO) algorithm [36] can be used to solve (10) for computational efficiency.

*D.2 Algorithm for solving the HOL optimization under the 0/1 loss*

Under the 0/1 loss, we have $w_{k,i} = 1$ if $k = Y_i^l - 1$ or $k = Y_i^r$, and $w_{k,i} = 0$ otherwise. Theorem 2 does not hold for the 0/1 loss. Thus, the optimization in (8) under the 0/1 loss becomes:

$$\min_{\eta, b_1, \dots, b_{K-1}} \sum_{k=1}^{K}\sum_{i\in D_k'} \max\left(0, 1 - Z_{k,i}(\eta^T\phi(\mathbf{x}_i) + b_k)\right) + \mu\|\eta\|_{\mathcal{H}}$$
$$\text{s.t. } b_1 \leq \cdots \leq b_{K-1}, \quad (12)$$

where $D_k$ is reduced to $D_k' = \{(\mathbf{x}_i, \mathcal{F}_i) | Y_i^r = k \text{ or } Y_i^l = k + 1; i = 1, \dots, n\}$ since $w_{k,i} = 0$ for samples in $D_k \backslash D_k'$.

To solve (12), we can adopt a similar approach as that for the MAE loss by first representing (12) into an equivalent form by introducing slack variables, and then solving the dual form of the primal problem. The details are skipped to save space.

## IV. EXPERIMENT

In this section, we compare the performance of HOL with existing OL models that can incorporate interval-labeled samples based on four public benchmarking datasets.

### A. Benchmarking datasets

**1) Abalones dataset**. This dataset is from the UCI repository. The goal is to predict the ages of abalones based on 8 features such as sex, length, diameter, and shell weight. The age of an abalone is determined by its number of rings. An abalone is considered 'young', 'adult', or 'old' (i.e., class 1, 2, or 3) if its number of rings falls in (0,5], [8,11] or [14,+∞), respectively. There is ambiguity in terms of class membership for abalones whose numbers of rings fall in [6,7] or [12,13]. These abalones are considered as samples with interval labels. Overall, this dataset includes 3057 samples with precise labels and 1120 samples with interval labels. The training set includes 500 precisely-labeled samples and 1120 interval-labeled samples. 2557 precisely-labeled samples are included in the test set. Note that in all the benchmarking experiments, the test set only includes samples with precise samples because we want to compute the accuracy of each model for classifying samples into their belonging ordinal classes.

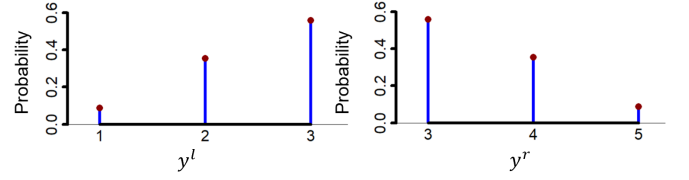

Figure 1: Discrete distributions to generate the lower bound $y^l$ (left) and upper bound $y^r$ (right) for a sample included in the training set.

**2) Auto-MPG dataset**. This dataset is from the UCI repository. The goal is to predict city-cycle fuel consumption in miles per gallon (mpg) based on 8 features such as car name, number of cylinders, horsepower, and weight. The mpg is divided into four ordinal classes: class 1-4 corresponds to mpg falling into (9-18], (18-23], (23-31], and (31-47]. This dataset only includes 398 automobile records with precise labels. To compose a training set that includes both precisely- and interval-labeled samples, we use a simulation method described below. Using the simulation, the training set includes 45 precisely-labeled samples and 55 interval-labeled samples on average over different simulation runs. 298 precisely-labeled samples were included in the test set.

**3) Boston house price dataset**. This dataset is from the UCI repository. The goal is to predict the median value of owner-occupied homes (medv) in $1000's for the Boston metropolitan area based on 13 socio-economic features such as the average room number, full-value property-tax rate, and accessibility to radial highways. The medv is divided into four ordinal classes: class 1-4 corresponds to medv falling into (5,16], (16,20], (20,25], and (25, 50]. This dataset only includes census tracts with 506 precise labels. To compose the training set that includes both precisely- and interval-labeled samples, we used a simulation method described below. Using the simulation, the training set includes 89 precisely-labeled samples and 111 interval-labeled samples on average over different simulation runs. 306 precisely-labeled samples were included in the test set.

**4) Eucalyptus soil conservation dataset.** This dataset is extracted from a published agricultural study in New Zealand (Thomson and McQueen 1996). The goal is to predict soil conservation capability of eucalyptus seedlots based on 19 features such as measurement of height, diameter by height, survival, and other contributing factors. The soil conservation capability is categorized into five ordinal classes: none, low, average, good, and best. This dataset only includes 736 eucalyptus seedlots with precise labels. Using the simulation described below, the training set includes 208 precisely-labeled samples and 292 interval-labeled samples on average over different simulation runs. 236 precisely-labeled samples were included in the test set.

Simulation method for training data generation: The last three datasets only contain precisely-labeled samples. To generate a training set with both precisely- and interval-labeled samples, we adopt the following procedure. First, among all the samples in the dataset, we split out a test set. For each remaining sample, the upper bound $y^r$ and lower bound $y^l$ for the interval label of this sample are generated as follows: $y^r$ is generated from a discrete distribution defined on $[y, K]$, where $y$ is the true label of the sample and $K$ is the total number of classes in the dataset. The probability mass function (PMF) of the discrete distribution is $\frac{1}{s}\{F(y^r + 1.5; y, 1) - F(y^r + 0.5; y, 1)\}$ for $y^r = y, y + 1, \ldots, K$. Here, $F(a; \mu, \sigma^2)$ denotes the cumulative density function of a normal distribution; $s$ is a normalizing factor to make sure that the PMFs over different values of $y^r$ sum up to be one. Similarly, the low bound $y^l$ is generated from another discrete distribution defined on $[1, y]$ with PMF being $\frac{1}{s}\{F(y^l + 1.5, 1) - F(y^l + 0.5; y, 1)\}$ for $y^l = 1, \ldots, y$. To better illustrate the two aforementioned discrete distributions, we give an example. Suppose there are five classes in a dataset, $K = 5$. Consider one sample from the dataset with true label $y = 3$. Fig. 1 shows the discrete distributions for the low bound $y^l$ and upper bound $y^r$ of the interval label. According to these distributions, it is possible that the generated $y^l$ and $y^r$ are the same, i.e., $y^l = y^r = y = 3$. In this case, the sample has a precise label. It is also possible that $y^l$ and $y^r$ are different, i.e., they can form any of the intervals such as [1,3], [2,3], [1,4], [2,4], [3,4], [1,5], [2,5] and [3,5]. Thus, this simulation method can generate a training set with both precisely- and interval-labeled samples to suit our need.

### B. Competing methods

We compare HOL with four competing methods:

<u>HOL/no-interval</u>: This is a special case of HOL where the model training only uses precisely-labeled samples and has to discard interval-labeled samples. Comparing HOL with this special case aims to demonstrate the benefit of including interval-labeled samples in training.

<u>HOL/mid-interval</u>: For interval-labeled samples, one could assign each sample to the middle class within the interval, creating a pseudo-precise label for the sample. These pseudo-precisely-labeled samples are then combined with precisely-labeled samples to train a HOL/no-interval model, referred to as the HOL/mid-interval method. It is worth noting that if the interval of a sample contains an even number of labels, there is no middle class. In this case, we randomly select from the middle two classes as the pseudo-precise label for the sample.

<u>VILMA</u>: **V**-shaped **i**nterval insensitive **l**oss **m**inimization **a**lgorithm (Antoniuk et al. 2016). This is a linear ordinal learning model that can integrate both precisely- and interval-labeled samples.

<u>PRIL</u>: **p**erceptron **r**anking using **i**nterval **l**abeled data (Manwani, 2018). This is a non-linear ordinal learning model that can integrate both precisely- and interval-labeled samples. As mentioned in Introduction, this method used an online learning algorithm for parameter estimation. However, the estimated discriminative ordinal functions in the form of combinations of kernel functions are rough as the coefficients of kernels can only take integer values.

### C. Model training and performance comparison

There are different variants of HOL, such as using the MAE loss or the 0/1 loss, choice of the regularization parameter $\lambda$ in the dual HOL optimization, and choice of the kernel function.

TABLE I
PERFORMANCE COMPARISON OF HOL AND COMPETING METHODS ON BENCHMARKING DATASETS (MEAN (STD) OVER 30 RUNS)

| Dataset | Method | Overall ACC ↑ | Ave. class-wise ACC ↑ | MSPE ↓ |
|---|---|---|---|---|
| Auto-MPG | HOL | **0.73** (0.038) | **0.74** (0.032) | **0.28** (0.039) |
|  | HOL/mid-interval | 0.71 (0.046) | 0.71 (0.05) | 0.30 (0.053) |
|  | HOL/no-interval | 0.71 (0.039) | 0.72 (0.033) | 0.30 (0.040) |
|  | PRIL | 0.70 (0.046) | 0.71 (0.048) | 0.30 (0.048) |
|  | VILMA | 0.66 (0.042) | 0.67 (0.043) | 0.36 (0.050) |
| Boston Housing | HOL | **0.72** (0.021) | **0.72** (0.020) | **0.30** (0.024) |
|  | HOL/mid-interval | 0.67 (0.038) | 0.66 (0.041) | 0.35 (0.038) |
|  | HOL/no-interval | 0.69 (0.036) | 0.68 (0.041) | 0.34 (0.040) |
|  | PRIL | 0.69 (0.060) | 0.68 (0.065) | 0.34 (0.071) |
|  | VILMA | 0.61 (0.060) | 0.61 (0.067) | 0.43 (0.072) |
| Eucalyptus | HOL | **0.67** (0.034) | **0.65** (0.032) | **0.35** (0.033) |
|  | HOL/mid-interval | 0.64 (0.039) | 0.62 (0.038) | 0.37 (0.042) |
|  | HOL/no-interval | 0.60 (0.028) | 0.61 (0.028) | 0.43 (0.031) |
|  | PRIL | 0.63 (0.078) | 0.63 (0.042) | 0.40 (0.098) |
|  | VILMA | 0.59 (0.036) | 0.60 (0.064) | 0.43 (0.040) |
| Abalones | HOL | **0.80** (0.035) | 0.77 (0.024) | **0.21** (0.035) |
|  | HOL/mid-interval | 0.50 (0.028) | 0.74 (0.012) | 0.51 (0.028) |
|  | HOL/no-interval | 0.79 (0.047) | **0.78** (0.029) | 0.22 (0.047) |
|  | PRIL | 0.72 (0.10) | 0.77 (0.026) | 0.28 (0.10) |
|  | VILMA | 0.66 (0.14) | 0.71 (0.12) | 0.34 (0.14) |

We consider these variants as tuning parameters and select the best combination of the tuning parameters by a grid search to minimize the classification error by HOL based on 10-fold cross validation (CV). The tuning parameters of the other methods are selected in a similar way. Then, the trained model under the best tuning parameters is used to classify samples in the test set. The following test metrics are computed and compared between different methods:

<u>Overall classification accuracy (overall ACC)</u>: This is the proportion of test samples that are correctly classified into their belonging classes.

<u>Average class-wise accuracy (ave. class-wise ACC)</u>: We computed the classification accuracy of each ordinal class and averaged over the class-wise accuracies.

<u>Mean absolute prediction error (MSPE)</u>: Since ordinal classes have a natural order, we computed the deviation of the predicted class with respect to the true class of each test sample and averaged over the deviations.

*D. Results*

Table I shows the overall ACC, average class-wise ACC, and MSPE of different methods on four benchmarking datasets. To help better visualize the results, Fig. 2 plots these performance metrics. In general, HOL outperforms HOL/no-interval and HOL/mid-interval. For the Abalones dataset, HOL/no-interval shows a slightly higher average class-wise ACC than HOL (0.78 compared to 0.77). This is because the Abalones dataset contains many precisely-labeled samples in the training set. Thus, whether or not the interval-labeled samples are included in training does not have a great impact.

Comparing between HOL/no-interval and HOL/mid-interval, the following observations can be drawn regarding their relative performance. When HOL/no-interval performs poorly, as seen in the Eucalyptus dataset with an overall ACC of only 0.60, adding pseudo-precisely-labeled samples by the HOL/mid-interval method results in an improved ACC of 0.64. Conversely, when HOL/no-interval performs well, such as for the Auto-MPG and Boston Housing datasets, incorporating pseudo-precisely-labeled samples does not provide significant benefit. Interestingly, HOL/mid-interval performs poorly in the Abalones dataset compared to HOL/no-interval. As mentioned earlier, the Abalones dataset has many precisely-labeled samples and utilizing them alone by HOL/no-interval already yields good performance. Also, there are numerous interval-labeled samples in this dataset. Creating pseudo-precise labels for these samples introduces a significant amount of noise in the model training, which degrades the performance of HOL/mid-interval. Furthermore, we were interested in exploring the performance of existing OL methods applied to the same datasets created for HOL/mid-interval, which include both precisely-labeled and pseudo-precisely labeled samples. We included popular OL methods such as ordinal random forest and ordinal logistic regression in this experiment. The ranges of ACC by these methods for the four benchmarking datasets are 0.62-0.71, 0.59-0.67, 0.59-0.65 and 0.47-0.62, respectively. These metrics are in similar ranges as HOL/mid-interval and worse than HOL.

Additionally, we can see that HOL outperforms the other competing methods, PRIL and VILMA. Even HOL/no-interval works better than or comparable to PRIL and VILMA in some metrics. This is because VILMA is a linear model which cannot capture the complicated non-linear relationships in these datasets. While PRIL is a non-linear model, it lacks a rigorous solution procedure and the algorithm cannot converge in some cases, which affects its performance.

## V. CASE STUDY ON EARLY PREDICTION OF AD

AD is a devastating neurodegenerative disease that currently affects 6.5 million people aged 65 and older in the U.S. [37], and over 55 million people worldwide [38]. There is currently no cure for AD. Early prediction bears the best hope for interventions to slow down the disease progression. MCI is a precursor to AD. Individuals with MCI show noticeable signs of memory loss and cognitive declines, but these symptoms are not severe enough to interfere their independent living. People with MCI progress to AD dementia in different speeds. Also, some people may not eventually progress to AD dementia if the underlying cause of their MCI symptoms is not related to AD but some other diseases. If one could accurately predict the speed of progressing to AD for each MCI subject (e.g., 'very fast', 'fast', 'moderate', or 'slow'), this capability will facilitate the development of more individually-optimized interventional strategies.

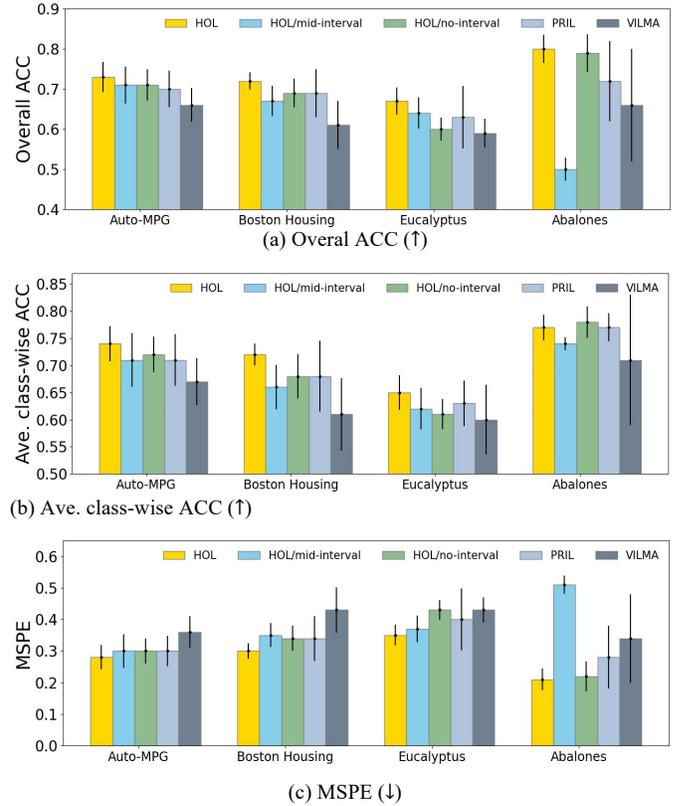

Figure 2: Bar charts for the performance metrics reported in Table I

In this section, we apply HOL to predicting the speed of MCI progression to AD in four ordinal classes, i.e., progressing to AD in one year (class 1), between year one and year two (class 2), between year two and year five (class 3), and beyond five years (class 4), using multi-modality neuroimaging and demographic/clinical datasets from the Alzheimer' Disease Neuroimaging Initiative (ADNI) project.

*A. Dataset description and preprocessing*

Introduction to ADNI: ADNI (http://adni.loni.ucla.edu) was launched in 2003 by the NIH, FDA, private pharmaceutical companies, and nonprofit organizations, as a $60 000 000, 5-year public-private partnership. After the initial ADNI project ended, subsequent efforts known as ADNI-GO, ADNI-2 and ADNI -3 added additional participants to augment the cohort. The primary goal of ADNI has been to test whether MRI, PET, other biological markers, and clinical and neuropsychological assessment can be combined to measure the progression of MCI and early AD. Determination of sensitive and specific markers of very early AD progression is intended to aid researchers and clinicians to develop new treatments and monitor their effectiveness, as well as lessen the time and cost of clinical trials. The Principal Investigator of this initiative is Michael W. Weiner, MD, VA Medical Center and University of California-

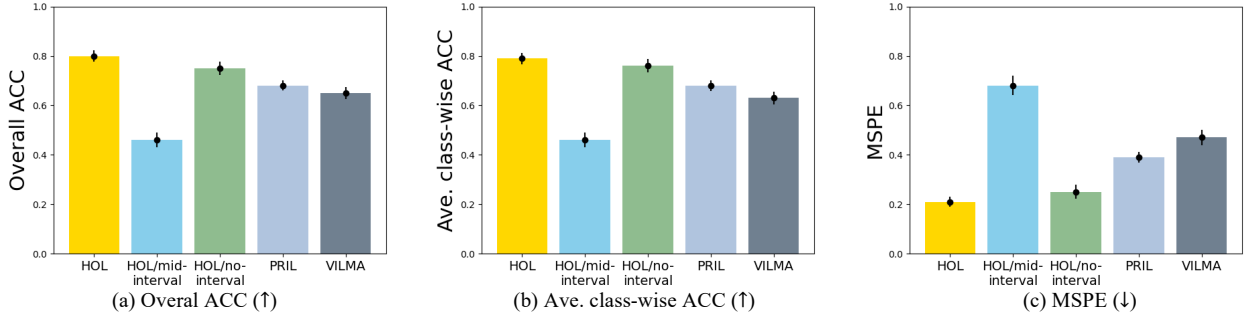

Figure 3: Bar charts for the performance metrics reported in Table II

San Francisco. ADNI is the result of efforts of many co-investigators from a broad range of academic institutions and private corporations, and subjects have been recruited from over 50 sites across the US and Canada. For up-to-date information, see http://www.adni-info.org/.

Patient inclusion: Our study includes 349 samples of 249 MCI patients from ADNI. Each sample corresponds to one visit record of a patient when neuroimaging and clinical data is collected. These visit records satisfy the following two conditions: at the time of visit, the patient (1) has MCI but not AD (since we want to predict the progression of MCI to AD), and (2) is "amyloid-positive" determined through a global standardized uptake value ratio (SUVR) cutoff of 1.4 which is validated by autopsy correlation with Thal amyloid phase [39]. In this study, we focus on patients with amyloid-positive MCI, because this sub-population is known to have an elevated risk of progressing to AD. Prediction of the speed of progressing for this sub-population has important clinical utility [40], [41]. Among the 349 samples, 204 have precise labels (55, 49, 54, 46 in class 1-4, respectively), whereas 145 have interval labels (55 samples in class 2, 3, or 4; 90 samples in class 3 or 4).

Neuroimage processing and feature computation: To train HOL and competing methods, we use T1 MRI, florbetapir-PET (a type of amyloid-PET), and demographic/clinical features of the patients. MRI is pre-processed using the FreeSurfer 7.1 software following standard procedures [42]. 156 commonly-used MRI features were extracted, including volumetric measures for 68 cortical regions of interest (ROIs), 14 sub-cortical structures, and 6 ventricle structures, as well as cortical thickness measures for the 68 ROIs. Amyloid-PET is pre-processed using a PET Unified Pipeline [43], [44] to obtain regional standardized uptake value ratios (SUVR) measurements for 68 cortical ROIs, 14 sub-cortical structures, 68 white matter structures, and a mean cortical feature —a total of 151 features. In addition, we also included demographic/clinical features including gender, age and education level; scores from common clinical instrument such as the Mini-Mental State Examination (MMSE) and the Clinical Dementia Rating Scale (CDR); status of the ε4 allele of apolipoprotein E (APOE) gene—a genetic risk factor of AD.

Feature screening: This dataset contains high-dimensional features compared to the benchmarking datasets. We use feature screening to reduce redundant features by adopting recursive feature elimination (RFE) [45]. In each iteration, RFE excludes the feature with the least importance based on a pre-defined metric. Two common metrics are adopted, one based on partial correlation between features [46] and the other based on multiple linear regression with each feature as the response and all other features as regressors [47]. Features with maximum partial correlation greater than 0.95 or with mean relative absolute error in the regression smaller than 0.05 are removed. The same feature screening procedure is applied to all the methods for fair comparison.

TABLE II
PERFORMANCE COMPARISON OF HOL AND COMPETING METHODS ON THE ADNI DATASET (MEAN (STD) OVER 30 RUNS)

| Methods | Overall ACC ↑ | Ave. class-wise ACC ↑ | MSPE ↓ |
|---|---|---|---|
| HOL | **0.80** (0.022) | **0.79** (0.023) | **0.21** (0.020) |
| HOL/mid-interval | 0.46 (0.030) | 0.46 (0.030) | 0.68 (0.039) |
| HOL/no-interval | 0.75 (0.028) | 0.76 (0.027) | 0.25 (0.028) |
| PRIL | 0.68 (0.020) | 0.68 (0.021) | 0.39 (0.022) |
| VILMA | 0.65 (0.025) | 0.63 (0.026) | 0.47 (0.030) |

### B. Modeling and results by different methods

We apply HOL and competing methods to this dataset. Specifically, we use a 30-fold cross validation (CV) scheme in which the patients are divided into 30 folds. One fold is left out as the validation set, whereas the other folds form the training set. Note that we divide patients not samples. This is to avoid including samples from the same patient into both the training and validation sets to cause overfitting. HOL is trained using the training set and the model is then applied to classify precisely-labeled samples in the validation set. This process is iterated over the 30 folds. The same 30-fold CV scheme is used for PRIL and VILMA. HOL/no-interval employs the same scheme except that the training set only includes precisely-labeled samples. For HOL/mid-interval the training set includes precisely-labeled samples and pseudo-precisely-labeled samples.

Table II shows the overall ACC, average class-wise ACC, and MSPE of different methods based on CV. To help better visualize the results, Fig. 3 plots these performance metrics. It is clear from the results that HOL outperforms the other methods across all metrics. HOL/ no-interval is the second-best method, even outperform PRIL and VILMA which can utilize interval-labeled samples in training. This can be attributed to the inherent limitations of these methods such as linearity and lack of a rigorous solution procedure.

HOL/mid-interval performs poorly, likely due to the pseudo-precise labels created from interval labels, which introduce a significant amount of noise into the training set. The dataset used in this case study includes a similar number of interval-labeled samples as precisely-labeled samples, meaning that almost half of the training data bears the risk of being wrongly labeled by using the pseudo labels. Looking across the results from the benchmarking datasets and from this case study, we found that HOL/mid-interval's performance can be quite unstable, depending on the dataset. Furthermore, we noted that the standard deviation of HOL/mid-interval is the largest compared to the other methods, further indicating the instability of this method. HOL/no-interval also has a large standard deviation, because it only uses precisely-labeled samples in training, whereas the other methods can additionally leverage interval-labeled samples to increase the training size.

## VI. Conclusions

We developed a new method, HOL, to integrate samples with both precise and interval labels to train a robust OL model. A non-linear model formulation was proposed to address application domains with complex data relationships. To tackle the challenge in model estimation, we further proposed a novel conversion method that converts the HOL formulation into an equivalent formulation of learning a set of binary classifiers with coupled parameters. The performance of HOL was demonstrated using benchmarking datasets and a real-world case study of predicting the speed of MCI progression to AD. HOL outperformed competing methods in these experiments.

This study has several limitations, which also inspire future research. First, the development of more efficient optimization solvers can be explored. In this study, a quadratic programing solver was used to solve the dual form of the HOL optimization when the sample size is relatively small and the SMO algorithm was used when the sample size is relatively large. For even larger data sets, more advanced algorithms are needed to solve HOL for computational efficiency. Second, we only discussed the algorithms under two loss functions, the MAE and 0/1 loss. Other types of losses can be explored in the HOL framework, such as the Mean Squared Error (MSE) and Huber Loss. Third, in the AD application, we used pre-extracted features from neuroimages to train HOL. Deep learning methods can be developed to take 3D neuroimages as input while at the same time leveraging precise- and interval-labeled samples in training. Last but not least, it would be interesting to apply HOL to other health care applications such as diagnosis of different grades of a disease and also to other diseases not AD.

## Acknowledgment

This research was supported by NIH grant 2R42AG053149-02A1 and NSF grant DMS-2053170. This research was also supported by NIH grants R01AG069453 and P30AG072980, the State of Arizona, and Banner Alzheimer's Foundation. Data collection and sharing for this project was funded by the Alzheimer's Disease Neuroimaging Initiative (ADNI) (National Institutes of Health Grant U01 AG024904) and DOD ADNI (Department of Defense award number W81XWH-12-2-0012). ADNI is funded by the National Institute on Aging, the National Institute of Biomedical Imaging and Bioengineering, and through generous contributions from the following: AbbVie, Alzheimer's Association; Alzheimer's Drug Discovery Foundation; Araclon Biotech; BioClinica, Inc.; Biogen; Bristol-Myers Squibb Company; CereSpir, Inc.; Cogstate; Eisai Inc.; Elan Pharmaceuticals, Inc.; Eli Lilly and Company; EuroImmun; F. Hoffmann-La Roche Ltd and its affiliated company Genentech, Inc.; Fujirebio; GE Healthcare; IXICO Ltd.;Janssen Alzheimer Immunotherapy Research & Development, LLC.; Johnson & Johnson Pharmaceutical Research & Development LLC.; Lumosity; Lundbeck; Merck & Co., Inc.;Meso Scale Diagnostics, LLC.; NeuroRx Research; Neurotrack Technologies; Novartis Pharmaceuticals Corporation; Pfizer Inc.; Piramal Imaging; Servier; Takeda Pharmaceutical Company; and Transition Therapeutics. The Canadian Institutes of Health Research is providing funds to support ADNI clinical sites in Canada. Private sector contributions are facilitated by the Foundation for the National Institutes of Health (www.fnih.org). The grantee organization is the Northern California Institute for Research and Education, and the study is coordinated by the Alzheimer's Therapeutic Research Institute at the University of Southern California. ADNI data are disseminated by the Laboratory for Neuro Imaging at the University of Southern California.

## Appendix:

### A. Proof of Theorem 1

For notation simplicity, denote $J(\mathbf{x}_i)$ by $J_i$ hereafter. First, we write out the loss function in (2) as follows:

$$L(J_i, \mathcal{F}_i) = \begin{cases} 0 & if J_i \in [Y_i^l, Y_i^r] \\ \sum_{k=J_i}^{Y_i^l-1} w_{k,i} & if\ J_i < Y_i^l \\ \sum_{k=Y_i^r}^{J_i-1} w_{k,i} & if\ J_i > Y_i^r \end{cases}$$

$$= \begin{cases} 0 & if\ J_i \in [Y_i^l, Y_i^r] \\ \sum_{k=1}^{Y_i^l-1} w_{k,i} I(k \geq J_i) & if\ J_i \leq Y_i^l - 1 \\ \sum_{k=Y_i^r}^{K-1} w_{k,i} I(k \leq J_i - 1) & if\ J_i > Y_i^r \end{cases}.$$

For any $k$, when $J_i \leq k$, we have $J_i = 1 + \sum_{j=1}^{K-1} I(f_j(\mathbf{x}_i) < 0) \leq k$, which indicates $f_k(\mathbf{x}_i) \geq 0$ due to the ordinal property of $f_j(\mathbf{x}_i)$ (i.e., $f_1(\mathbf{x}_i) \leq \cdots \leq f_{K-1}(\mathbf{x}_i)$); or else the inequality will not be satisfied. Thus, $I(k \geq J_i) = I(f_k(\mathbf{x}_i) \geq 0)$. Similarly, $I(k \leq J_i - 1) = I(f_k(\mathbf{x}_i) < 0)$. Thus, $L(J_i, \mathcal{F}_i)$ can be re-written as

$$L(J_i, \mathcal{F}_i) = \begin{cases} 0 & if\ J \in [Y_i^l, Y_i^r] \\ \sum_{k=1}^{Y_i^l - 1} w_{k,i} I(f_k(\mathbf{x}_i) \geq 0) & if\ f_{Y_i^l - 1}(\mathbf{x}_i) \geq 0. \\ \sum_{k=Y_i^r}^{K-1} w_{k,i} I(f_k(\mathbf{x}_i) < 0) & if\ f_{Y_i^r}(\mathbf{x}_i) < 0 \end{cases}$$

According to the definition of $Z_{k,i}$, it can be equivalently written as

$$Z_{k,i} = \begin{cases} 1, & if\ k \geq Y_i^r \\ -1, & if\ k < Y_i^l \\ 0, & if\ k \in [Y_i^l, Y_i^r) \end{cases}.$$

Then we have

$$L(J_i, \mathcal{F}_i) = \begin{cases} 0 & if\ J \in [Y_i^l, Y_i^r] \\ \sum_{k=1}^{Y_i^l - 1} w_{k,i} I(Z_{k,i} f_k(\mathbf{x}_i) < 0) & if\ Z_{Y_i^l - 1} f_{Y_i^l - 1}(\mathbf{x}_i) \leq 0 \\ \sum_{k=Y_i^r}^{K-1} w_{k,i} I(Z_{k,i} f_k(\mathbf{x}_i) < 0) & if\ Z_{Y_i^r} f_{Y_i^r}(\mathbf{x}_i) < 0 \end{cases}$$
$$= \sum_{k=1}^{Y_i^l - 1} w_{k,i} I(Z_{k,i} f_k(\mathbf{x}_i) < 0) + \sum_{k=Y_i^r}^{K-1} w_{k,i} I(Z_{k,i} f_k(\mathbf{x}_i) < 0)$$
$$= \sum_{k=1}^{K-1} w_{k,i} I(Z_{k,i} f_k(\mathbf{x}_i) < 0).$$

Hence,

$$L(J, \mathcal{F}) = \sum_{i=1}^n L(J_i, \mathcal{F}_i)$$
$$= \sum_{i=1}^n \sum_{k=1}^{K-1} w_{k,i} I(Z_{k,i} f_k(\mathbf{x}_i) < 0)$$
$$= \sum_{k=1}^{K-1} \sum_{i=1}^n w_{k,i} I(Z_{k,i} f_k(\mathbf{x}_i) < 0)$$
$$= \sum_{k=1}^{K-1} \sum_{i \in D_k} w_{k,i} I(Z_{k,i} f_k(\mathbf{x}_i) < 0).$$

The last equation is based on the definition of $Z_{k,i}$. Finally, according to the definition of $\mathcal{B}(f, Z)$, we have $L(J, \mathcal{F}) = \mathcal{B}(f, Z)$. ∎

*B. Proof of Theorem 2*

Let $\mathcal{B}_{MAE}(f, Z)$ be the optimization function in (9) (i.e., $\mathcal{B}_{MAE}(f, Z) = \sum_{k=1}^{K-1} \sum_{i \in D_k} \max(0, 1 - Z_{k,i}(\boldsymbol{\eta}^T \phi(\mathbf{x}_i) + b_k)) + \mu \|\boldsymbol{\eta}\|_{\mathcal{H}}$), and $\boldsymbol{\eta}, b_1, \ldots, b_{K-1}$ be its optimal solution. Due to convexity, the $b_k$'s are closed intervals, i.e., $b_k = [d_k, d_k'], k = 1, \ldots, K-1$. We will prove Theorem 2 by contradiction. Suppose constraint of $b_1 \leq \cdots \leq b_{K-1}$ in (8) does not automatically hold in the solution of (9). That is, assume $b_k > b_{k+1}$ for some $k$.

Let $k'$ be another rank index different from $k$. Let $D_{k'}^l$ denote the lower subset of $D_{k'}$ as

$$D_{k'}^l = \{(\mathbf{x}_i, \mathcal{F}_i) | \mathcal{F}_i \subseteq [1, k']; i = 1, \ldots, n\},$$

and $D_{k'}^r$ denote the upper subset of $D_{k'}$ as

$$D_{k'}^r = \{(\mathbf{x}_i, \mathcal{F}_i) | \mathcal{F}_i \subseteq [k' + 1, K]; i = 1, \ldots, n\}.$$

Let $I_{k'}^l(b_k) = \{i \in D_{k'}^l : \boldsymbol{\eta}^T \phi(\mathbf{x}_i) + b_k \leq 1\}$ be the subset of $D_{k'}^l$, where samples satisfy $\boldsymbol{\eta}^T \phi(\mathbf{x}_i) + b_k \leq 1$, and $I_{k'}^r(b_k) =$ $\{i \in D_{k'}^r : \boldsymbol{\eta}^T \phi(\mathbf{x}_i) + b_k \geq -1\}$ be the subset of $D_{k'}^r$, where samples satisfy $\boldsymbol{\eta}^T \phi(\mathbf{x}_i) + b_k \geq -1$.

Under the MAE loss, the subgradient of the loss function with respect to $b_k$ is

$$g_{k'}(b_k) = \frac{\partial \mathcal{B}(f, Z)}{\partial b_k} = -|I_{k'}^l(b_k)| + |I_{k'}^r(b_k)|.$$

Then we have

$g_{k+1}(b_{k+1}) - g_k(b_{k+1})$
$= -|I_{k+1}^l(b_{k+1})| + |I_{k+1}^r(b_{k+1})| + |I_k^l(b_{k+1})| - |I_k^r(b_{k+1})|$
$= (-|I_{k+1}^l(b_{k+1})| + |I_k^l(b_{k+1})|) + (|I_{k+1}^r(b_{k+1})| - |I_k^r(b_{k+1})|)$.

It is clear that $-|I_{k+1}^l(b_{k+1})| + |I_k^l(b_{k+1})| \leq 0$ and $|I_{k+1}^r(b_{k+1})| - |I_k^r(b_{k+1})| \leq 0$. We have $g_{k+1}(b_{k+1}) - g_k(b_{k+1}) \leq 0$. That is,

$$g_{k+1}(b_{k+1}) \leq g_k(b_{k+1}). \tag{13}$$

Since $b_{k+1}$ is an optimal solution, we know that $g_{k+1}(b_{k+1}) \geq 0$.

Due to $b_k > b_{k+1}$ and the convexity of $\mathcal{B}_{MAE}(f, Z)$ on $b_k, k = 1, \ldots, K - 1$, we have $g_k(b_{k+1}) < 0$. Then $g_k(b_{k+1}) < g_{k+1}(b_{k+1})$, which contradicts (13). ∎

*C. Proof of Theorem 3*

Let $\boldsymbol{\gamma} = (\{\alpha_{i \in D_1}^1\}, \ldots, \{\alpha_{i \in D_{K-1}}^{K-1}\})^T$, $v_i^k$ for $i$ and $k$, be the Lagrange multipliers, and $\lambda$ be a tuning parameter. The Lagrangian for the primal HOL optimization in (10) is

$$L = \frac{1}{2} \boldsymbol{\eta}^T \boldsymbol{\eta} - \sum_{k=1}^{K-1} \sum_{i \in D_k} \alpha_i^k \left( Z_{k,i} (\boldsymbol{\eta}^T \phi(\mathbf{x}_i) + b_k) - 1 + \xi_i^k \right) + \lambda \sum_{k=1}^{K-1} \sum_{i \in D_k} \xi_i^k - \sum_{k=1}^{K-1} \sum_{i \in D_k} v_i^k \xi_i^k. \tag{14}$$

Then the optimal solution of the primal problem in (10) is equivalent to the solution of the following optimization:

$$\max_{\boldsymbol{\gamma}, \boldsymbol{v}} \min_{\boldsymbol{\eta}, b, \boldsymbol{\xi}} L. \tag{15}$$

The KKT conditions for the primal problem require the following to hold:

$$\nabla_{\boldsymbol{\eta}} L = \boldsymbol{\eta} - \sum_{k=1}^{K-1} \sum_{i \in D_k} \alpha_i^k Z_{k,i} \phi(\mathbf{x}_i) = 0,$$
$$\nabla_{b_k} L = -\sum_{i \in D_k} \alpha_i^k Z_{k,i} = 0, k = 1, \ldots, K-1, \tag{16}$$
$$\nabla_{\xi_i^k} L = -\alpha_i^k + \lambda - v_i^k = 0, i \in D_k, k = 1, \ldots, K-1. \tag{17}$$

Then we have

$$\boldsymbol{\eta} = \sum_{k=1}^{K-1} \sum_{i \in D_k} \alpha_i^k \phi(\mathbf{x}_i), \tag{18}$$
$$v_i^k = -\alpha_i^k + \lambda, i \in D_k, k = 1, \ldots, K-1. \tag{19}$$

Inserting (18) and (19) into the optimization in (15), after simplification we can get

$$\max_{\boldsymbol{\gamma}, \mu} L$$

$$= \tfrac{1}{2}\boldsymbol{\eta}^T\boldsymbol{\eta} - \sum_{k=1}^{K-1}\sum_{i\in D_k} \alpha_i^k\bigl(Z_{k,i}(\boldsymbol{\eta}^T\phi(\mathbf{x}_i)+b_k) - 1 + \xi_i^k\bigr)$$
$$+\lambda\sum_{k=1}^{K-1}\sum_{i\in D_k}\xi_i^k - \sum_{k=1}^{K-1}\sum_{i\in D_k}(-\alpha_i^k + \lambda)\xi_i^k$$
$$= \tfrac{1}{2}\boldsymbol{\eta}^T\boldsymbol{\eta} - \sum_{k=1}^{K-1}\sum_{i\in D_k}\alpha_i^k\bigl(Z_{k,i}(\boldsymbol{\eta}^T\phi(\mathbf{x}_i)+b_k) - 1\bigr)$$
$$= \tfrac{1}{2}\boldsymbol{\eta}^T\boldsymbol{\eta} - \sum_{k=1}^{K-1}\sum_{i\in D_k}\boldsymbol{\eta}^T\alpha_i^k Z_{k,i}\phi(\mathbf{x}_i) -$$
$$\sum_{k=1}^{K-1} b_k \sum_{i\in D_k}\alpha_i^k Z_{k,i} + \sum_{k=1}^{K-1}\sum_{i\in D_k}\alpha_i^k.$$
$$= \tfrac{1}{2}\boldsymbol{\eta}^T\boldsymbol{\eta} - \sum_{k=1}^{K-1}\sum_{i\in D_k}\boldsymbol{\eta}^T\alpha_i^k Z_{k,i}\phi(\mathbf{x}_i) + \sum_{k=1}^{K-1}\sum_{i\in D_k}\alpha_i^k.$$
(20)

Then inserting (18) into the optimization in (20), we can have

$$\max_{\boldsymbol{\gamma}} L$$
$$= -\tfrac{1}{2}\Bigl(\sum_{k=1}^{K-1}\sum_{i\in D_k}\alpha_i^k Z_{k,i}\phi(\mathbf{x}_i)\Bigr)^T\Bigl(\sum_{k=1}^{K-1}\sum_{i\in D_k}\alpha_i^k Z_{k,i}\phi(\mathbf{x}_i)\Bigr) + \sum_{k=1}^{K-1}\sum_{i\in D_k}\alpha_i^k$$
$$= -\tfrac{1}{2}\boldsymbol{\gamma}^T \mathbf{Z}\mathbf{C}\mathbf{Z}\boldsymbol{\gamma} + \sum_{k=1}^{K-1}\sum_{i\in D_k}\alpha_i^k$$

Additionally, the conditions in (16) give rise to the constraints of $-\sum_{i\in D_k}\alpha_i^k Z_{k,i} = 0, k = 1, \dots, K-1$.

The conditions in (19) give rise to the constraints of

$$0 \le \alpha_i^k \le \lambda, i \in D_k, k = 1, \dots, K-1.$$

Finally, the dual problem becomes

$$\min_{\boldsymbol{\gamma}} \tfrac{1}{2}\boldsymbol{\gamma}^T \mathbf{Z}\mathbf{C}\mathbf{Z}\boldsymbol{\gamma} - \sum_{k=1}^{K-1}\sum_{i\in D_k}\alpha_i^k,$$
$$\text{s.t. } \sum_{i\in D_k}\alpha_i^k Z_{k,i} = 0, k = 1, \dots, K-1;$$
$$0 \le \alpha_i^k \le \lambda, i \in D_k, k = 1, \dots, K-1. \quad \blacksquare$$

*D. List of mathematical notations*

| Notation | Description |
|---|---|
| $\mathcal{X}$ | feature space |
| $\mathcal{Y}$ | label space |
| $f_k(\cdot)$ | ranking function at rank $k$ |
| $k$ | index to ranks |
| $I(\cdot)$ | indicator function |
| $J(\cdot)$ | label predicted function |
| $\phi(\cdot)$ | transformations of feature vectors |
| $\boldsymbol{\eta}$ | combination coefficients of transformed feature vectors |
| $b_k$ | intercepts of ranking function at rank $k$ |
| $i$ | index to samples |
| $\mathcal{F}_i$ | true label interval for sample $i$ |
| $Y_i^l$ | lower bound of label interval for sample $i$ |
| $Y_i^r$ | upper bound of label interval for sample $i$ |
| $L(\cdot)$ | loss function |
| $\|\cdot\|_{\mathcal{H}}$ | norm in a metric space $\mathcal{H}$ |
| $\mu$ | weight of model complexity |
| $J_i$ | predicted label for sample $i$ |
| $L_{MAE}(\cdot)$ | mean absolute error (MAE) loss function |
| $L_{0/1}(\cdot)$ | 0/1 loss function |
| $D_k$ | subset of the whole training set excluding samples whose label interval includes $k$ |
| $Z_{k,i}$ | $1, -1,$ or $0$ corresponds to $k \ge Y_i^r$, $k < Y_i^l$, or $k \in [Y_i^l, Y_i^r]$ |
| $\mathcal{B}(f,Z)$ | total loss of $K-1$ binary classifiers |
| $L(J,\mathcal{F})$ | hybrid ordinal learner (HOL) loss |
| $\xi_i^k$ | slack variable of $k^{th}$ ranking function for sample $i$ |
| $\lambda$ | tuning parameter |
| $\mathbf{Z}$ | diagonal matrix |
| $\mathbf{C}$ | covariance matrix |
| $k(\cdot)$ | kernel function defined on the feature space |
| $\boldsymbol{\gamma}$ | Lagrange multipliers |
| $D_k'$ | reduced form of $D_k$ under the 0/1 loss |
| $y^l$ | lower bound of interval label generated for the precisely-labeled sample |
| $y^r$ | upper bound of interval label generated for the precisely-labeled sample |
| $y$ | true label of the precisely-labeled sample |
| $K$ | total number of classes in the dataset |
| $F(a;\mu,\sigma^2)$ | cumulative density function of a normal distribution |
| $s$ | normalizing factor of probability mass function (PMF) |
| $d_k$ | lower bound of $b_k$ |
| $d_k'$ | upper bound of $b_k$ |
| $k'$ | another index to ranks |
| $D_{k'}^l$ | lower subset of $D_{k'}$, where sample label interval $\mathcal{F}_i \subseteq [1, k']$ |
| $D_{k'}^r$ | upper subset of $D_{k'}$, where sample label interval $\mathcal{F}_i \subseteq [k'+1, K]$ |
| $I_{k'}^l(b_k)$ | a subset of $D_{k'}^l$, where samples satisfy a specific condition involving $b_k$ |
| $I_{k'}^r(b_k)$ | a subset of $D_{k'}^r$, where samples satisfy a specific condition involving $b_k$ |
| $|\cdot|$ | sample size of input set |
| $v_i^k$ | Lagrange multiplier of $k^{th}$ ranking function for sample $i \in D_k$ |
| $\alpha_i^k$ | Lagrange multiplier of $k^{th}$ ranking function for sample $i \in D_k$ |

IEEE TRANSCATIONS ON AUTOMATION SCIENCE AND ENGINEERING, VOL. X, NO. X, XXX 2022    14

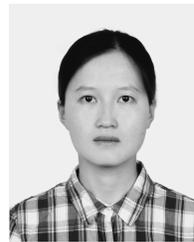

**Lujia Wang** received her B.S. degree in Mathematics and Applied Mathematics from Nankai University, Tianjin, China, in 2013, and an M.S. degree in Probability and Mathematical Statistics from Chinese Academy of Sciences, Beijing, China, in 2016. She received her Ph.D. in the School of Industrial and Systems Engineering, Georgia Institute of Technology, Atlanta, GA, USA. Her research interests include machine learning and biomedical imaging analytics.

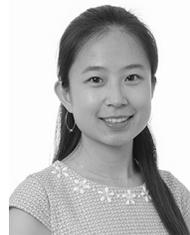

**Hairong Wang** received her B.A. degree in Mathematics from University of Oxford, Oxford, UK. She is a Ph.D. student in the School of Industrial and Systems Engineering, Georgia Institute of Technology, Atlanta, GA, USA. Her research interests are knowledge-informed machine learning, deep learning, and biomedical imaging analytics. She is a member of IISE, INFORMS, and IEEE.

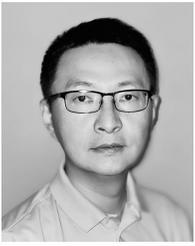
**Yi Su** received his Ph.D. degree in Biomedical Engineering from Mayo Graduate School at Rochester, MN. He directs the Computational Image Analysis Laboratory at the Banner Alzheimer's Institute in Phoenix, AZ. His research focuses on the development and application of novel imaging and data analytics techniques to better understand aging and neurodegenerative diseases such as Alzheimer's disease.

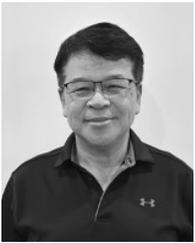
**Fleming Lure** received his Ph.D. degree in Electrical Engineering from the Pennsylvania State University, State College, PA. He is a Chief Product Officer in MS Technologies Corp, Rockville, MD, USA. His research interests are computer aided detection and machine learning for disease detection, diagnosis, and prognosis. He has led a team to receive the first FDA-approved early-stage lung cancer detection system on radiograph. He is a member of Radiological Society of North America (RSNA).

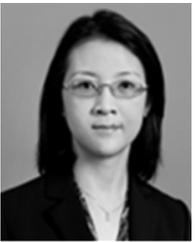
**Jing Li** received her Ph.D. degree in Industrial and Operations Engineering from the University of Michigan at Ann Arbor, MI. She is a Professor in the School of Industrial and Systems Engineering, Georgia Institute of Technology, Atlanta, GA, USA. Her research interests are statistical modeling and machine learning for health care applications. She is a recipient of NSF CAREER award. She is a member of IISE, INFORMS, and IEEE.